\documentclass[conference]{IEEEtran}
\IEEEoverridecommandlockouts
\usepackage{cite}
\usepackage{amsmath,amssymb,amsfonts}
\usepackage{graphicx}
\usepackage{textcomp}
\usepackage{xcolor}

\usepackage{todonotes}
\usepackage{url}
\usepackage{algorithm}
\usepackage{algorithmic}
\usepackage{eqparbox}
\usepackage{hyperref}

\DeclareMathOperator*{\argmin}{arg\,min}

\newcommand{\vk}[1]{{\color{violet}#1}}

\def\LossZB{\mathcal{L}_{\mathcal{Z}}^{I}}
\def\LossMB{\mathcal{L}_{\mathcal{Z}}^{II}}
\def\LossZCos{\mathcal{L}_{\mathcal{Z}}^{III}}
\def\LossZCosUn{\mathcal{L}_{\mathcal{Z}}^{IV}}

\def\LossZ{\mathcal{L}_{\mathcal{Z}}}
\def\LossX{\mathcal{L}_{\mathcal{X}}}
\def\LossW{\mathcal{L}_{\mathcal{W}}}

\def\BibTeX{{\rm B\kern-.05em{\sc i\kern-.025em b}\kern-.08em
    T\kern-.1667em\lower.7ex\hbox{E}\kern-.125emX}}
\begin{document}

\title{Evaluation of Security of ML-based Watermarking: Copy and Removal Attacks\\
\thanks{This research was partially funded by the SNF Sinergia project (CRSII5-193716): Robust deep density models for high-energy particle physics and solar flare analysis (RODEM).}
}

\author{\IEEEauthorblockN{Vitaliy Kinakh}
\IEEEauthorblockA{\textit{Department of Computer Science} \\
\textit{University of Geneva}\\
Geneva, Switzerland \\
vitaliy.kinakh@unige.ch}
\and
\IEEEauthorblockN{Brian Pulfer}
\IEEEauthorblockA{\textit{Department of Computer Science} \\
\textit{University of Geneva}\\
Geneva, Switzerland \\
brian.pulfer@unige.ch}
\and
\IEEEauthorblockN{Yury Belousov}
\IEEEauthorblockA{\textit{Department of Computer Science} \\
\textit{University of Geneva}\\
Geneva, Switzerland \\
yury.belousov@unige.ch}
\and
\IEEEauthorblockN{Pierre Fernandez}
\IEEEauthorblockA{\textit{Meta, FAIR} \\
\textit{University of Rennes, Inria, CNRS, IRISA}\\
pierre.fernandez@inria.fr}
\and
\IEEEauthorblockN{Teddy Furon}
\IEEEauthorblockA{\textit{University of Rennes, Inria, CNRS, IRISA} \\
Rennes, France \\
teddy.furon@inria.fr}
\and
\IEEEauthorblockN{Slava Voloshynovskiy}
\IEEEauthorblockA{\textit{Department of Computer Science} \\
\textit{University of Geneva}\\
Geneva, Switzerland \\
svolos@unige.ch}
}

\maketitle
\begin{abstract}
The vast amounts of digital content captured from the real world or AI-generated media necessitate methods for copyright protection, traceability, or data provenance verification. Digital watermarking serves as a crucial approach to address these challenges. Its evolution spans three generations: handcrafted, autoencoder-based, and foundation model based methods. While the robustness of these systems is well-documented, the security against adversarial attacks remains underexplored. This paper evaluates the security of foundation models' latent space digital watermarking systems that utilize adversarial embedding techniques. A series of experiments investigate the security dimensions under copy and removal attacks, providing empirical insights into these systems' vulnerabilities. All experimental codes and results are available in the \href{https://github.com/vkinakh/ssl-watermarking-attacks}{repository}.
\end{abstract}


\section{Introduction}
The emergence of a vast amount of content is reshaping our digital landscape. This content is either captured directly from the real world, i.e., physically produced, or created via digital algorithms, i.e., synthetically generated. This spans various media, including images, videos, audio, and text.

In this new landscape, verifying the integrity, authenticity, and provenance poses significant challenges to maintaining trust, preventing misinformation, preserving the integrity of legal evidence, and upholding ethical standards. 
Notably, the EU AI Act recognizes the risks linked with the recent machine learning (ML) models and the content they generate~\cite{EC_AI_Act}.

Digital watermarking is a crucial technical means in copyright protection and traceability. This technology aims to meet four primary requirements: imperceptibility, payload, robustness and security. While its robustness is well-documented, the security aspects, particularly of recent schemes based on ML, remain underexplored.

Foundation Models (FMs) and, notably, Vision Foundation Models (VFMs) are central to this evolving digital ecosystem~\cite{oquab2023dinov2, radford2021learningclip}. 
They represent a significant advancement in ML capabilities. These large pre-trained neural networks, refined on extensive and diverse datasets, are versatile tools. Many downstream applications use VFMs for analyzing content, like image classification, semantic segmentation, object detection, content retrieval, and tracking.

Based on this idea, a similar trend in watermarking~\cite{vukotic2020classification, fernandez2022watermarking} aims to leverage the robustness and performance of these models. They usually utilize adversarial embedding techniques to hide information in VFMs' latent spaces. It makes the resulting watermarking robust and very versatile: able to operate on images with different resolutions, with a variable payload and a manually defined trade-off between robustness and quality. This paper evaluates and highlights the brittle security of these methods. Addressing this gap enhances the understanding and development of secure digital watermarking in our increasingly digital world.

The main contributions are as follows: a) We introduce two classes of attacks against latent space watermarking, specifically focusing on copy and removal attacks; b) We investigate the performance of these attacks on a state-of-the-art technique within this class of watermarking, evaluating both zero-bit and multi-bit watermarking schemes; c) We demonstrate the impact of target selection strategies in the effectiveness of removal attacks; d) We provide a comprehensive analysis of the vulnerability of DINOv1~\cite{caron2021emerging}, highlighting the necessity for future research on a broader range of foundation models.
\section{State of the Art of Watermarking}
{\bf Digital watermarking} embeds information within digital media, balancing (1) {\em imperceptibility} - the distortion induced by the watermark is not perceptible for a human observer, (2) {\em payload} - the amount of data embedded in the content, (3) {\em robustness} - the ability to retrieve the hidden message under a given set of distortions and (4) {\em security} - the ability to withstand attacks exploiting the system's vulnerability. Techniques vary from {\em zero-bit watermarking}, where a mark is embedded into a content using a secret key and the detection assesses the presence of this mark within the content, to {\em multi-bit watermarking}, which encodes a message in content, and the decoder retrieves the embedded message bit by bit.

Digital watermarking has evolved across three generations  differentiated by their embedding domains:

\begin{enumerate} \addtolength{\itemsep}{-0.1\baselineskip}
    \item \textbf{\(\mathcal{DW}_1\)}: Techniques in this category embed watermarks in the {\em spatial} or {\em transform} domains, including DFT \cite{urvoy2014perceptual, Voloshynovskiy:1997:MVD}, DCT \cite{bors1996image, pereira2000effective}, Fourier-Melline \cite{pereira1999template}, and DWT \cite{xia1998wavelet} domains, with both {\em zero-bit}  \cite{furon2007constructive} and {\em  multi-bit} watermarking \cite{hernandez1998performance, Voloshynovskiy:2001:MDW}. These methods aim for invisibility and basic robustness, employing additive or quantization-based embedding techniques \cite{chen2001quantization, eggers2001quantization}.
 \item \textbf{\(\mathcal{DW}_2\)}: This group jointly trains ML-based encoder and decoder for adaptive embedding \cite{kandi2017exploring, lee2020convolutional, zhu2018hidden}, focusing on content-driven robustness enhancements. These methods involve training under differentiable distortions, including adversarial settings \cite{luo2020distortion, wen2019romark}, and require adaptation to new types of datasets and distortions.
 \item \textbf{\(\mathcal{DW}_3\)}: The most recent advancement explores watermarking by using iterative adversarial-like embeddings in the latent spaces of pre-trained models, either trained on a supervised task~\cite{vukotic2020classification} or with VFMs~\cite{fernandez2022watermarking}. In this paper, we consider DINOv1 model \cite{caron2021emerging}. DINOv1 is a self-supervised learning computer vision model, that uses student-teacher framework, the student predicts teacher's output for different image augmentations. DINOv1 captures semantic information and performs well on tasks like image classification and object detection.
\end{enumerate}

{\bf Security of digital watermarking}: Extensive robustness and security assessments have been conducted on the $\mathcal{DW}_1$ group. These studies pinpoint the difficulty to fight against the {\em copy attack}~\cite{Kutter:2000:WCA}, the {\em remodulation attack}~\cite{voloshynovskiy2001attacks}, and the {\em sensitivity attack}  \cite{linnartz1998analysis, earl2007tangential, comesana2006blind, el2007sensitivity}. Conversely, the exploration of the security of $\mathcal{DW}_2$ and $\mathcal{DW}_3$ watermarking in the face of adversarial attacks is still in its infancy. This early inquiry phase highlights a significant gap in our understanding of their security, indicating a critical field for research endeavours.

{\bf Notations}: We denote by $\mathcal{X} =\mathbb{R}^{H \times W \times C}$ the space of images of size ${H \times W \times C}$. A trained VFM is denoted as $f_\phi:\mathcal{X} \to \mathcal{Z}$ mapping the image space to the latent space $\mathcal{Z} =\mathbb{R}^d$. Notations $\mathbf{x}_0$, $\mathbf{x}_w$, and $\mathbf{x}_a$ stand for the original, watermarked and attacked images in $\mathcal{X}$, $\mathbf{z}_0$, $\mathbf{z}_w$ and $\mathbf{z}_a$ correspond to their latent space representations in $\mathcal{Z}$.  We have $\mathbf{x}_w=w(\mathbf{x}_0,m,k)$ where $m$ is the message to be hidden and $k$ the secret key, and $\mathbf{x}_a=t(\mathbf{x}_w)$ where $t$ is an image transformation 
pertaining to a set of attacks $\mathcal{T}$.

The distortion is measured by $\LossX : {\mathcal X} \times {\mathcal X} \rightarrow \mathbb{R}^+$. In the case of mean square error (MSE), $\LossX\left(\mathbf{x}_0, \mathbf{x}_w\right) = || \mathbf{x}_0 - \mathbf{x}_w||_2^2/H/W/C \leq D_w$, where $D_w$  defines the embedding distortion budget between the original and watermarked images. If the size and geometry of the image after the attack are preserved, one can also define the attack distortion $\LossX\left(\mathbf{x}_w, \mathbf{x}_a\right)$. The MSE is usually given in log scale by the peak signal-to-noise ratio $\text{PSNR}_w = 10\log_{10} \left( 255^2/\LossX\left(\mathbf{x}_0, \mathbf{x}_w\right) \right) $ for measuring quality of watermarked imaged and $\text{PSNR}_a = 10\log_{10} \left( 255^2/\LossX\left(\mathbf{x}_w, \mathbf{x}_a\right) \right)$ for attacked images. 
\section{VFM-based Adversarial Embedding Watermarking}

This section summarizes the watermarking method~\cite{fernandez2022watermarking} by first accounting for the detection/decoding stage.

\subsection{Detection and Decoding}
We consider two scenarios: zero-bit (detection only) and multi-bit watermarking (decoding the hidden message).

\textbf{Zero-Bit.} Given a secret carrier ${\bf w} \in {\mathcal Z}$ s.t. $\|{\bf w}\| = 1$, generated from the secret key $k$, that represents a 0-bit watermarking, the detection region is the dual hypercone:
\begin{equation}
\label{eq:detector}
{\mathcal D}_k := \{{\bf z} \in \mathbb{R}^d : |{\bf z}^T{\bf w}| > \|{\bf z\| \cos(\gamma)\}}.
\end{equation}
The angle $\gamma$ is defined by the targeted false acceptance rate $P^t_{\text{fa}}$, that is theoretically given for a non-watermarked $\bf x$ as:
\begin{equation}
P^t_{\text{fa}} := \mathbb{P}\left[f_\phi({\bf x}) \in {\mathcal D}_K | {K} \sim \mathcal{U})\right] = 1 - I_{\cos^2(\gamma)}\left(\frac{1}{2}, \frac{d-1}{2}\right),
\end{equation}
where $I_{\tau}(\alpha, \beta)$ is the regularized Beta incomplete function.
The following function gauges how ${\bf z}$ is close \vk{to} ${\mathcal D}_k$:
\begin{equation}
\LossZB({\bf z},{\bf w}) =  \|{\bf z}\|^2 \cos^2 (\theta) - ({\bf z}^T{\bf w})^2.
\label{eq:Loss_ZeroBit}
\end{equation}
Its sign indicates whether ${\bf z}$ lies inside ${\mathcal D}_k$, its amplitude indicates how far ${\bf z}$ is from ${\mathcal D}_k$ or deep inside ${\mathcal D}_k$.

\textbf{Multi-Bit.} The hidden message is ${m} = (m_1, \ldots, m_\ell) \in \{-1, 1\}^\ell$. The random generator seeded with the secret key $k$ produces an orthogonal family of carriers $\{{\bf w}_1, \ldots, {\bf w}_\ell\} \subset \mathcal{Z}$. The decoder retrieves $\hat{{m}}$ as the sign of the projections:
\begin{equation*}
\hat{{m}}=\left(\operatorname{sign}\left(f_\phi({\bf x})^{\top} {\bf w}_1\right), \ldots, \operatorname{sign}\left(f_\phi({\bf x})^{\top} {\bf w}_\ell\right)\right).
\end{equation*}
The following function gauges how $\mathbf{z}$ lies deep inside the decoding region within a margin \(\mu \geq 0\) on the projections.
\begin{equation}
\LossMB({\bf z},m)=\frac{1}{\ell} \sum_{i=1}^\ell \max \left(0, \mu-\left({\bf z}^{\top} {\bf w}_i\right) \cdot m_i\right).
\label{eq:Loss_MultiBit}
\end{equation}

\subsection{Watermark embedding}\label{sec:wm_embedding}

The embedding takes an original image ${\bf x}_0 \in \mathcal{X}$ and outputs a visually similar image ${\bf x}_w \in \mathcal{X}$. The previous section defines a loss function $\LossZ$ in the latent space, be it~\eqref{eq:Loss_ZeroBit} or~\eqref{eq:Loss_MultiBit}. The embedding aims at minimizing this loss under the constraint of distortion defined in the image domain. Augmentations are introduced to make the watermark signal more robust.
These are image modifications belonging to a set ${\mathcal T}$ of typical attacks with a range of parameters, such as rotation, crops and blur.
The application of attack $t\in\mathcal T$ to image ${\bf x}$ writes as $t(\mathbf{x})\in\mathcal{X}$.

The losses $\LossZ$ and $\LossX$ are combined as follows:
\begin{equation}
\LossW({\bf x}, {\bf x}_0, t) := \lambda \LossZ(f_\phi(t({\bf x}))) + \LossX({\bf x}, {\bf x}_0),
\end{equation}
where $\lambda$ controls the trade-off between two terms:
$\LossZ$ aims to push the feature of any transformation of ${\bf x}_w$  deep inside the detection/decoding region, while $\LossX$ favors low distortion.
The embedding is typical from the adversarial ML literature minimizing an Expectation over Transformation (EoT)~\cite{athalye2018synthesizing}:
\begin{equation}
{\bf x}_w := \argmin_{{\bf x} \in C({\bf x}_0)} \mathbb{E}_{T \sim \mathcal{U}(\mathcal{T})}[\LossW({\bf x}, {\bf x}_0, T)],
\label{eq:embedding}
\end{equation}
where $C({\bf x}_0) \subset {\mathcal X}$ is the set of admissible images w.r.t. the original one. It is defined by two steps of normalization applied to the pixel-wise difference ${\boldsymbol \delta}_{0} = {\bf x} - {\bf x}_0$: (1) we apply a SSIM~\cite{wang2004image} heatmap attenuation, which scales ${\boldsymbol \delta}_{0}$ pixel-wise to hide the information in perceptually less visible areas of the image; (2) we set a target PSNR and rescale ${\boldsymbol \delta}_{0}$ accordingly. 
\section{Attacks against ML-based digital watermarking}

This paper assumes the attacker knows neither the secret key $k$ nor the message \(m\). However, the main brick of the system is the foundation model $f_\phi$ which is open-sourced and therefore a white-box for the attacker. 

\begin{figure}[t]
\vspace{0 cm}
\centering
	\includegraphics[width=8.9 cm]{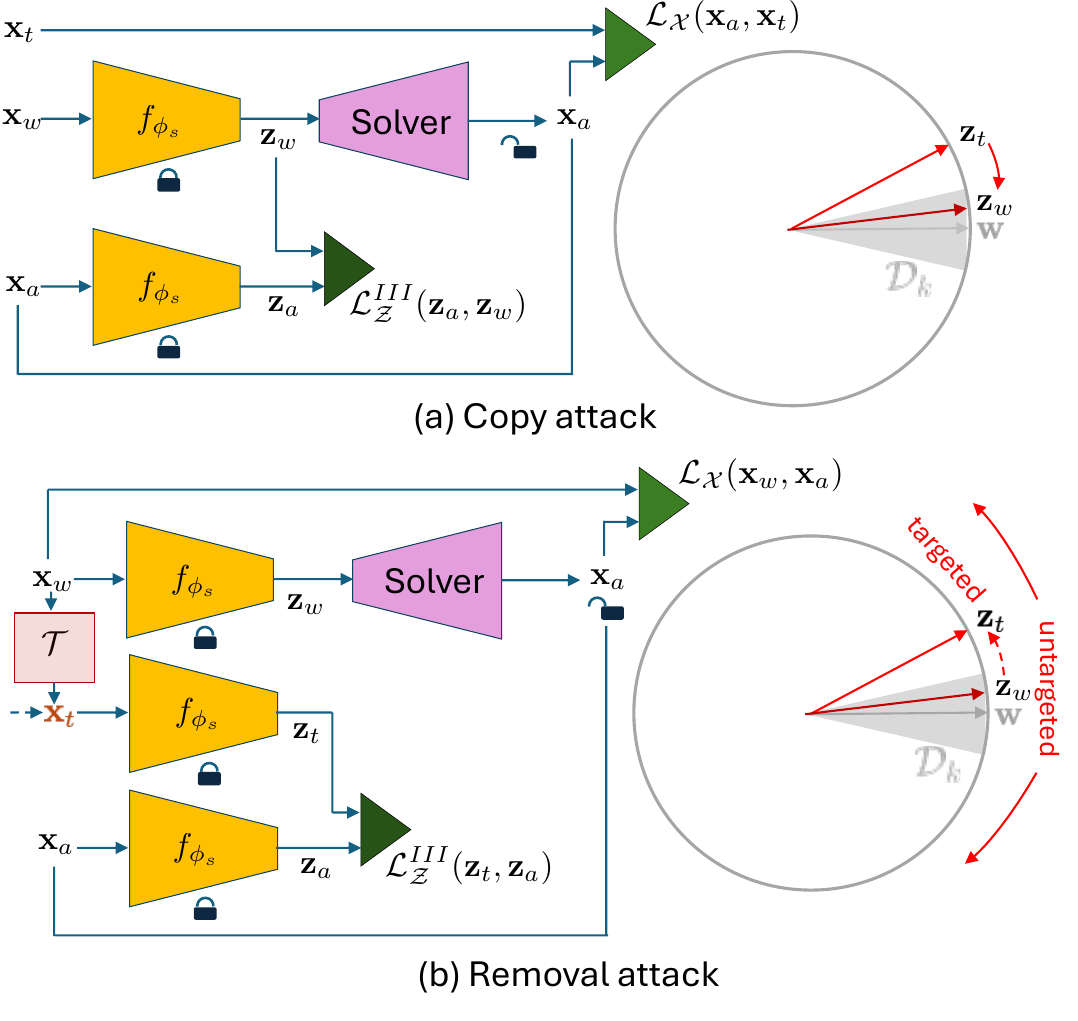}
        \vspace{-0.5cm}	
	\caption{Generalized diagram explaining the proposed (a) copy and (b) untargted and targeted removal attacks (on the example of zero-bit watermarking in the latent space). The secret carrier $\bf w$ and the decision region ${\mathcal D}_k$ (show in gray) are unknown for the attacker.}
\label{fig:DWM_attack}	
\end{figure}

\subsection{Watermark Copy Attack}

The objective of a \emph{copy attack} is to maximize the probability of falsely accepting a non-watermarked image as a watermarked one. Given a watermarked image $\mathbf{x}_w$ and a target image $\mathbf{x}_t$, the attack seeks to transfer the watermark from $\mathbf{x}_w$ to $\mathbf{x}_t$ without knowledge of the message $m$ or the key $k$.

In contrast to the traditional copy attack~\cite{Kutter:2000:WCA}, Fig.~\ref{fig:DWM_attack}a proposes a generalization  across various embedding domains that does not necessitate the additivity of the embedding. 

Given the watermarked image $\mathbf{x}_w$ and the target image $\mathbf{x}_t$, our copy attack generates an attacked image $\mathbf{x}_a$ that is perceptually close to  $\mathbf{x}_t$ according to the loss function $\LossX(\mathbf{x}_t, \mathbf{x}_a)$. Concurrently, the latent representation $\mathbf{z}_a$ of the attacked image is driven towards the latent representation $\mathbf{z}_w$ of the watermarked image as per a loss function $\LossZCos(\mathbf{z}_a, \mathbf{z}_w)$.
The total loss for the generalized copy attack is formulated as:
\begin{equation}
\label{eq:copy_attack}
\mathcal{L}^{\text{C}}_{\mathcal{A}}(\mathbf{x}_a, \mathbf{x}_w, \mathbf{x}_t) = \LossX(\mathbf{x}_a, \mathbf{x}_t) + \lambda \LossZCos(\mathbf{z}_a, \mathbf{z}_w),
\end{equation}
where $\lambda$ is a weighting factor that balances the contributions of the perceptual and latent similarity terms. The latent space loss is defined as $\LossZCos({\bf z}_a, {\bf z}_w) = - \frac{\mathbf{z}_a^T \mathbf{z}_w}{\sqrt{\left\|\mathbf{z}_a\right\|_2^2\|\mathbf{z}_w\|_2^2}}$ for both zero-bit and multi-bit watermarking. Minimization is achieved via gradient descent over $N$ iterations. Similar to the watermark embedding~\eqref{eq:embedding}, the attack also involves two normalization steps applied to the difference $\boldsymbol{\delta}_{at} = \mathbf{x}_a - \mathbf{x}_t$, i.e. the SSIM masking and the rescaling to impose a certain ${\text{PSNR}}_a$. The final image is rounded to quantized pixels. The algorithm of the proposed copy attack is presented below.

\begin{algorithm}
\caption{Copy Attack}
\label{alg:copy_attack}
\begin{algorithmic}[1]
\STATE \textbf{Input}: $\mathbf{x}_w$: watermarked image, $\mathbf{x}_t$: target image; $f_\phi$: feature extractor (FM)
\STATE $\mathbf{z}_w \leftarrow f_\phi(\mathbf{x}_w)$, $\mathbf{x}_a \leftarrow \mathbf{x}_t$ \COMMENT{ // initialize}
\FOR{$t = 0, \ldots, N-1$}
    \STATE $\mathbf{x}_a \stackrel{\text{constraints}}{\longleftarrow} \mathbf{x}_a$ \COMMENT{// impose constraints via ${\boldsymbol \delta}_{at}$}
    \STATE $\mathbf{z}_a \leftarrow f_\phi(\mathbf{x}_a)$ \COMMENT{// compute latent representation}
    \STATE $\mathbf{x}_a \leftarrow \mathbf{x}_a + \eta \times \operatorname{Adam}\left(\mathcal{L}^{\text{C}}_{\mathcal{A}}(\mathbf{x}_a, \mathbf{x}_w, \mathbf{x}_t)\right)$\\ \COMMENT{// update the image}
\ENDFOR
    \STATE $\mathbf{x}_a \stackrel{\text{constraints}}{\longleftarrow} \mathbf{x}_a$ \COMMENT{// impose constraints via ${\boldsymbol \delta}_{at}$, rounding}
\STATE \textbf{Return}: Attacked image $\mathbf{x}_a$\\
\end{algorithmic}
\end{algorithm}

{\bf Extension to multiple watermarked images}. When multiple images $\{ {{\bf x}_w}_n \}_{n=1}^L$ watermarked with the same key and the same message (in the case of multi-bit watermarking) are available to the attacker, one can compensate the lack of knowledge of the acceptance region $\mathcal D$ by solving the following optimization problem: for ${\mathbf{z}_w}_n = f_{\phi}( {{\bf x}_w}_n ),\,\forall n\in[L]$, 
\begin{equation}
\label{eq:multiple_copy}
\mathcal{L}_{\mathcal{A}}^{\mathrm{C}}(\mathbf{x}_a, \mathbf{x}_w, \mathbf{x}_t) = \LossX(\mathbf{x}_a, \mathbf{x}_t) +
\frac{\lambda}{L} \sum_{n=1}^L \LossZCos\left(\mathbf{z}_a, {{\bf z}_w}_n \right).
\end{equation}

In our experiments, we observe the very high success rates of the targeted attacks in the setup where $L=1$. Thus, we do not experiment with these attacks in Sec.~\ref{sec:expes}. 

\subsection{Watermark Removal Attack}

The watermark removal damages the watermarked image to maximize the probability of miss detection (zero-bit watermarking), or the bit error rate (BER) (multi-bit watermarking).
 
Our proposal is to jeopardize the latent space representation with the hope of diminishing the presence of the watermark. Specifically, given a watermarked image $\mathbf{x}_w$, the attack generates an attacked image $\mathbf{x}_a$ perceptually similar to $\mathbf{x}_w$ while ensuring that its latent representation $\mathbf{z}_a$ is far from $\mathbf{z}_w$. This strategy does not require an additive approximation of the embedding. Neither the watermark detector/decoder output nor the secret key $k$ is required.

Technically, the watermark removal can be achieved by a) {\em untargeted attack} (removal-untargeted, R-U) or b) {\em targeted  attack} (R-T). In the untargeted case, the loss function is defined $\LossZCosUn({\bf z}_a, {\bf z}_w) = \frac{(\mathbf{z}_a^T \mathbf{z}_w) ^ 2}{\sqrt{\left\|\mathbf{z}_a\right\|_2^2\|\mathbf{z}_w\|_2^2}}$ for both zero-bit and multi-bit watermarking.
\begin{equation}
\label{eq:removal_utargeted}
\mathcal{L}_{\mathcal{A}}^{\mathrm{R-U}}(\mathbf{x}_w, \mathbf{x}_a) = \LossX(\mathbf{x}_w, \mathbf{x}_a) - \lambda \LossZCosUn({\bf z}_w, {\bf z}_a).
\end{equation}
The targeted removal attack generates an attacked image $\mathbf{x}_a$ that is perceptually close to the watermarked image $\mathbf{x}_w$ while its latent representation $\mathbf{z}_a$ gets away from $\mathbf{w}$ and instead aligns with the latent representation of a target image $\mathbf{z}_t$:
\begin{equation}
\label{eq:removal}
\mathcal{L}_{\mathcal{A}}^{\mathrm{R-T}}(\mathbf{x}_w, \mathbf{x}_t, \mathbf{x}_a) = \LossX(\mathbf{x}_w, \mathbf{x}_a) + \lambda \LossZCos({\bf z}_t, {\bf z}_a),
\end{equation}
Minimization of the total loss is achieved via stochastic gradient descent over $N$ iterations. The final image is obtained with the SSIM masking and scaling of the perturbation ${\boldsymbol \delta}_{aw} = \mathbf{x}_a - \mathbf{x}_w$ to achieve a given ${\text{PSNR}}_a$, and rounding. 

\begin{algorithm}
\caption{Watermark Removal Attack}
\label{alg:watermark_removal_attack}
\begin{algorithmic}[1]
\STATE \textbf{Input}: $\mathbf{x}_w$: watermarked image, $\mathbf{x}_t$: target image; $f_\phi$: feature extractor (FM), $attack\_type$: type of attack (targeted or untargeted)
\STATE \textbf{Compute}: $\mathbf{z}_t = f_\phi(\mathbf{x}_t)$
\STATE \textbf{Initialize}: $\mathbf{x}_a \leftarrow \mathbf{x}_w$
\FOR{$t = 0, \ldots, N-1$}
    \STATE $\mathbf{x}_a \stackrel{\text{constraints}}{\longleftarrow} \mathbf{x}_a$ 
    \COMMENT{// impose constraints via ${\boldsymbol \delta}_{aw}$}
    \STATE $\mathbf{z}_a \leftarrow f_\phi(\mathbf{x}_a)$ \COMMENT{// compute latent representation}
    \IF{$attack\_type$ == ``untargeted''}
        \STATE $\mathbf{x}_a \leftarrow \mathbf{x}_a + \eta \times \operatorname{Adam}(\mathcal{L}_{\mathcal{A}}^{\mathrm{R-U}}(\mathbf{x}_w,  \mathbf{x}_a))$ \\ \COMMENT{// update the image according to untargeted attack}
    \ELSIF{$attack\_type$ == ``targeted''}
        \STATE $\mathbf{x}_a \leftarrow \mathbf{x}_a + \eta \times \operatorname{Adam}(\mathcal{L}_{\mathcal{A}}^{\mathrm{R-T}}(\mathbf{x}_w, \mathbf{x}_t, \mathbf{x}_a))$ \\ \COMMENT{// update the image according to targeted attack}
    \ENDIF
\ENDFOR
\STATE $\mathbf{x}_a \stackrel{\text{constraints}}{\longleftarrow} \mathbf{x}_a$ \COMMENT{// impose constraints via ${\boldsymbol \delta}_{aw}$, rounding}
\STATE \textbf{Return}: Attacked image $\mathbf{x}_a$
\end{algorithmic}
\end{algorithm}

{\bf 
The target selection} during the removal attack plays an important role for the success of the attack.  Three strategies are being considered. 1) Choosing any random non-watermarked image ${\bf x}_t$. 2) Setting target to be a heavily degraded version of ${\bf x}_w$ for which the watermark is no longer detected. Then, the optimization~\eqref{eq:removal} restores a better image quality. 3) Selecting random watermarking carrier as the new target.
\section{Experimental Results}\label{sec:expes}
The implementation of the studied zero-bit and multi-bit watermarking is based on the paper~\cite{fernandez2022watermarking}. The ResNet-50 trained with DINOv1~\cite{caron2021emerging} is used as the vision backbone. All experiments are performed on the DIV2K dataset~\cite{agustsson2017ntire} with typical image size $2000 \times 1500$.  Unless specified otherwise, the experiments are repeated using 10 different keys for watermark embedding and detection on a subset of 800 images from DIV2K. In all experiments, the $\text{PSNR}_w$ of the original watermarked image is fixed at 42 dB, and the target $\text{PSNR}_a$ varies from 30 to 45~dB. For most of the attacks, the actually achieved $\text{PSNR}_a$ is higher than the above target value.

\subsection{Investigation on the Copy Attack}
The first experiment investigates the robustness against the copy attack. 
The goal is to copy the watermark on un-watermarked images from a single watermarked image. The $\text{PSNR}_w$ of the original watermarked image is fixed at 42 dB.

For zero-bit watermarking, the attack success rate measures the proportion of crafted images that are wrongly flagged by the watermark detection~\eqref{eq:detector}, for different targeted probabilities of false acceptance $P^t_{\text{fa}}\in\{10^{-5}, 10^{-6}, 10^{-7}\}$.  The optimization of Alg.~\ref{alg:copy_attack} achieves the attack success rate equals one for the entire range of studied $\text{PSNR}_a$ and targeted false acceptance $P^t_{\text{fa}}$. This confirms the strength of the copy attack.

\begin{figure}[b]
    \centering
    \includegraphics[width=0.8\linewidth, clip, trim=0 0cm 0 0]{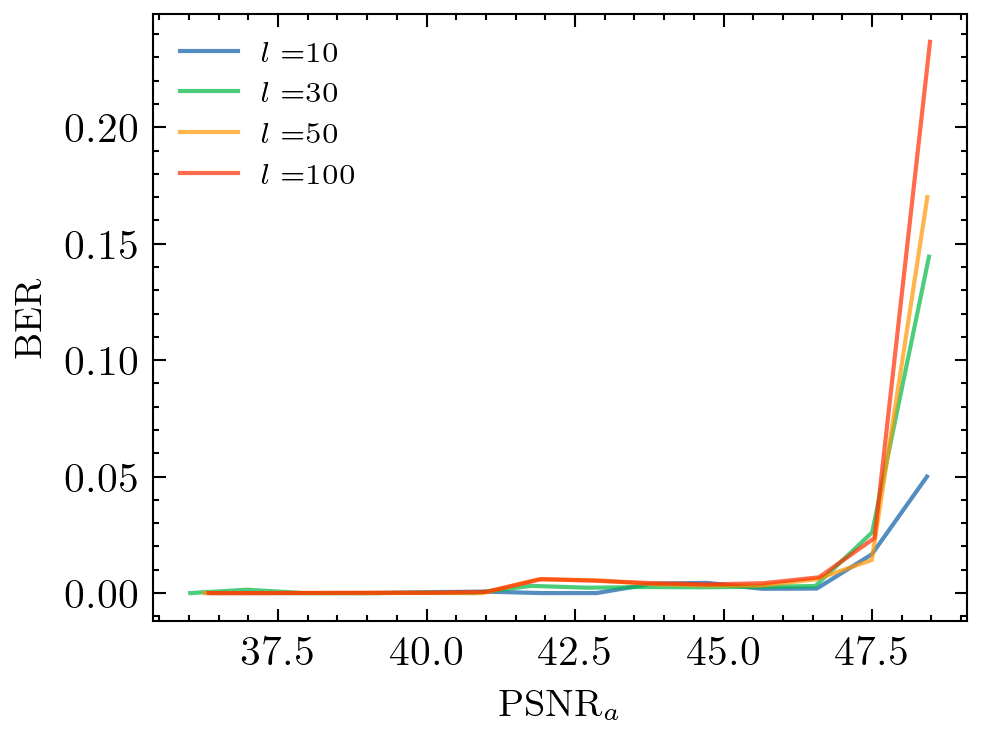}
    \vspace{-0.5cm}	
    \caption{Bit Error Rate (BER) for multi-bit watermarking under the copy attack with varying $\text{PSNR}_a$ and watermark payloads $\ell$. The attack can successfully copy the binary message (BER $<$ 1\%) of the watermarked image into any non-watermarked image, even at very low distortion budgets ($\text{PSNR}_a = 45$~dB).}
    \label{fig:copy_attack_multi_bit}
\end{figure}

The second experiment involves multi-bit watermarking. The watermark payload varies $\ell\in\{10, 30, 50, 100\}$ bits. Fig.~\ref{fig:copy_attack_multi_bit} shows that, at low values of $\text{PSNR}_a$ (strong attack distortions), the multi-bit watermarks are perfectly copied. At higher values of $\text{PSNR}_a$ (weak attack), the BER naturally increases but not significantly. The increase of message length causes higher value of BER obtained at high $\text{PSNR}_a =$ 47.5~dB, but for lower $\text{PSNR}_a$ the impact of watermark payload length is insignificant. This demonstrates strong clonability. 

\subsection{Investigation on the Removal Attack}

This section studies both untargeted and targeted removal attacks against zero-bit and multi-bit watermarking. In contrast to the copy attack, the attack success rate now measures the probability of miss $P_{\text m}$ for zero-bit watermarking, \textit{i.e.}, the proportion of watermarked images that are no longer detected after the attack, and the BER for multi-bit watermarking. 

\begin{figure}[t]
    \centering
    \includegraphics[width=0.8\linewidth]{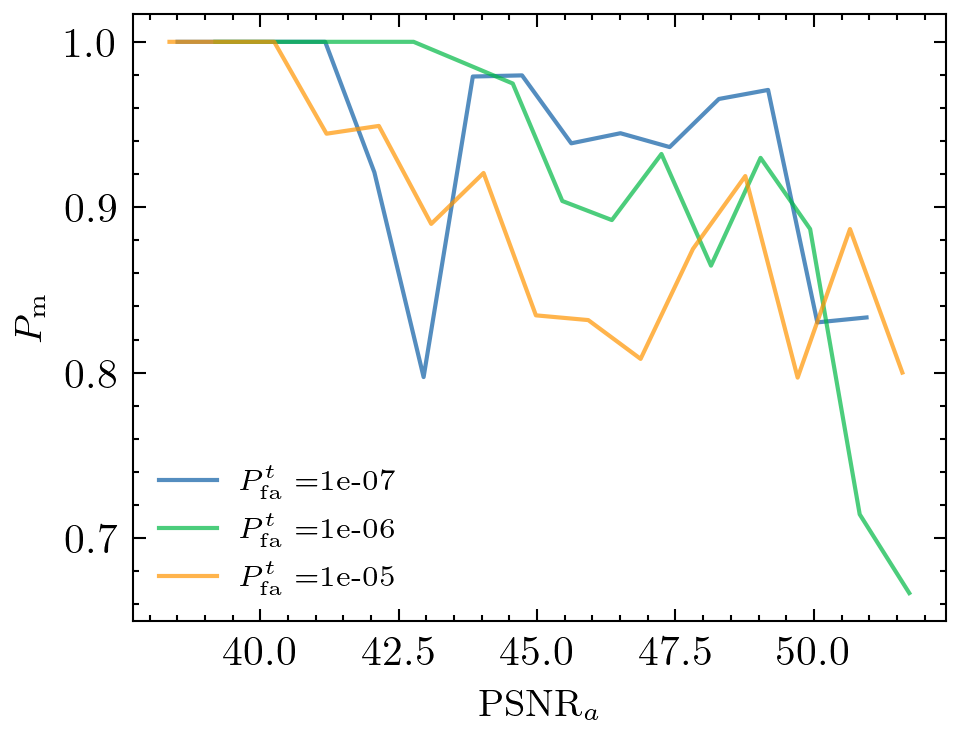}
    \vspace{-0.5cm}
    \caption{
    Probability of miss for zero-bit watermarking under untargeted removal attack against $\text{PSNR}_a$ of the attacked image, for varying probability of false acceptance. The untargeted attack achieves $P_{\text m}$ close to 1 at lower values of $\text{PSNR}_a$ around 40~dB, while $P_{\text m}$ decreases with the increase of $\text{PSNR}_a$ towards 50~dB.}
    \label{fig:removal_unt_attack_zero_bit}
\end{figure}

\begin{figure}[b]
    \centering
    \includegraphics[width=0.8\linewidth]{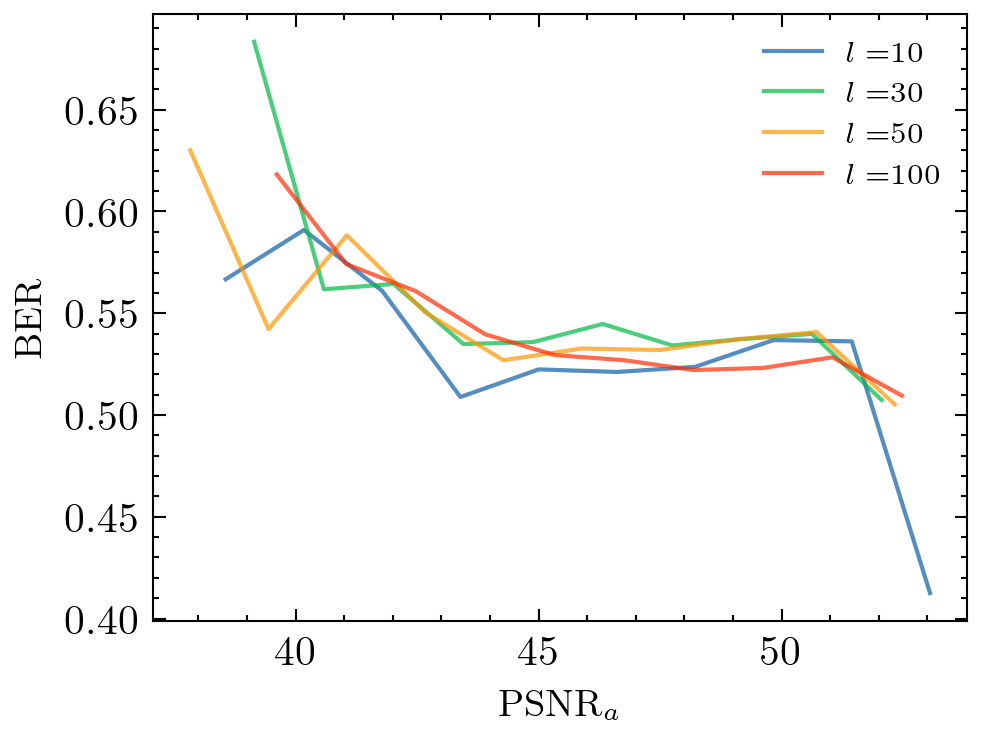}
    \vspace{-0.5cm}
    \caption{Bit Error Rate for multi-bit watermarking under untargeted removal attack against $\text{PSNR}_a$ at varying payload of $\ell$ bits. The attack increases the BER significantly, inverting the majority of the hidden bits.}
\label{fig:removal_unt_attack_multi_bit}
\end{figure}

The untargeted removal attack~\eqref{eq:removal_utargeted} does not require any target. Fig.~\ref{fig:removal_unt_attack_zero_bit} reports the observed $P_{\text m}$ for the zero-bit watermarking detection at different targeted probabilities of false acceptance.  On the other hand, Fig.~\ref{fig:removal_unt_attack_multi_bit} shows the influence on BER for multi-bit watermarking.  The untargeted removal attack significantly impacts the performance of both watermarking schemes. 

In contrast to the untargeted removal attack, the targeted removal attack needs to select the target ${\bf x}_t$ and accordingly ${\bf z}_t = f_{\phi}({{\bf x}_t})$. The target image selection strategies include random selection of  ${\bf x}_t$ denoted as ``other image'', selecting the denoised watermark image as ${\bf x}_t = d_{\text{Wiener}}({\bf x}_w)$, and selecting directly ${\bf z}_t$ randomly in the latent space.

\begin{figure}[t]
    \centering
    \includegraphics[width=0.8\linewidth]{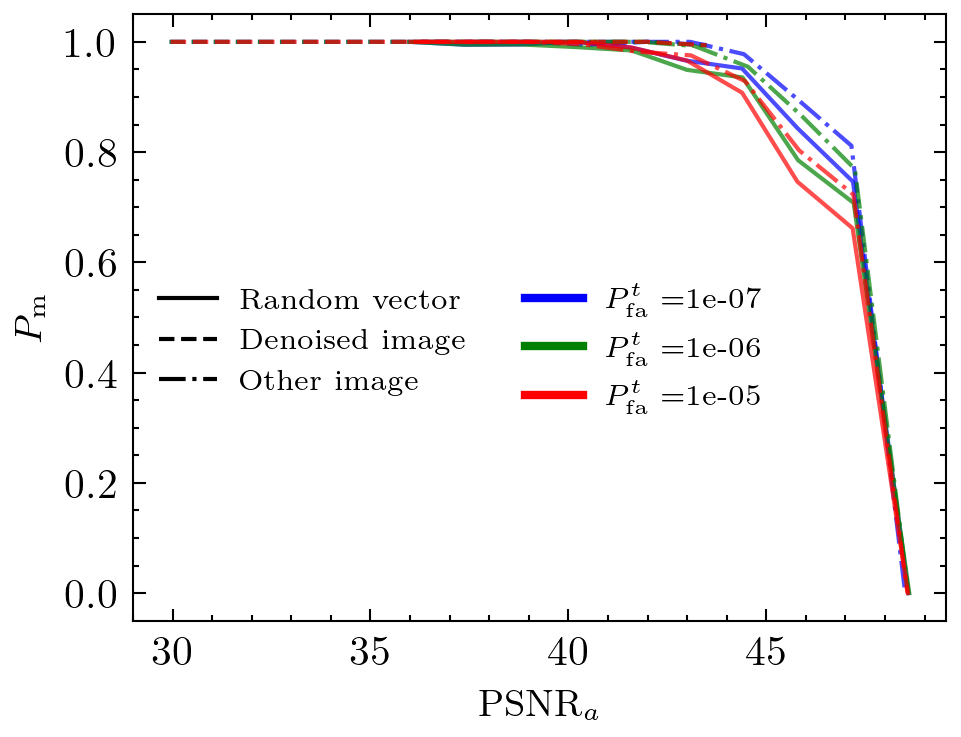}
    \caption{
    Probability of miss for zero-bit watermarking under targeted removal attack with different target image selection strategies. All kinds of targeted attacks achieve better success rates than the untargeted ones.}
    \label{fig:removal_attack_zero_bit}
\end{figure}

Fig.~\ref{fig:removal_attack_zero_bit} shows the $P_{\text m}$ under targeted removal attack for zero-bit watermarking with the required target probability of false acceptance: $10^{-5}$, $10^{-6}$ and $10^{-7}$. The selection of the denoised image based on Wiener filter with size $25 \times 25$ as a target image provides the best results in maximization of probability of miss for all values of probability of false acceptance. Comparing the results from Fig.~\ref{fig:removal_attack_zero_bit} and Fig.~\ref{fig:removal_unt_attack_zero_bit}, one can conclude that both untargeted and targeted removal attacks achieve $P_{\text m}$ close to 1,  for $\text{PSNR}_a \leq$  41~dB, that demonstrates high efficiency of both strategies.

\begin{figure}[b]
    \centering
    \includegraphics[width=0.8\linewidth]{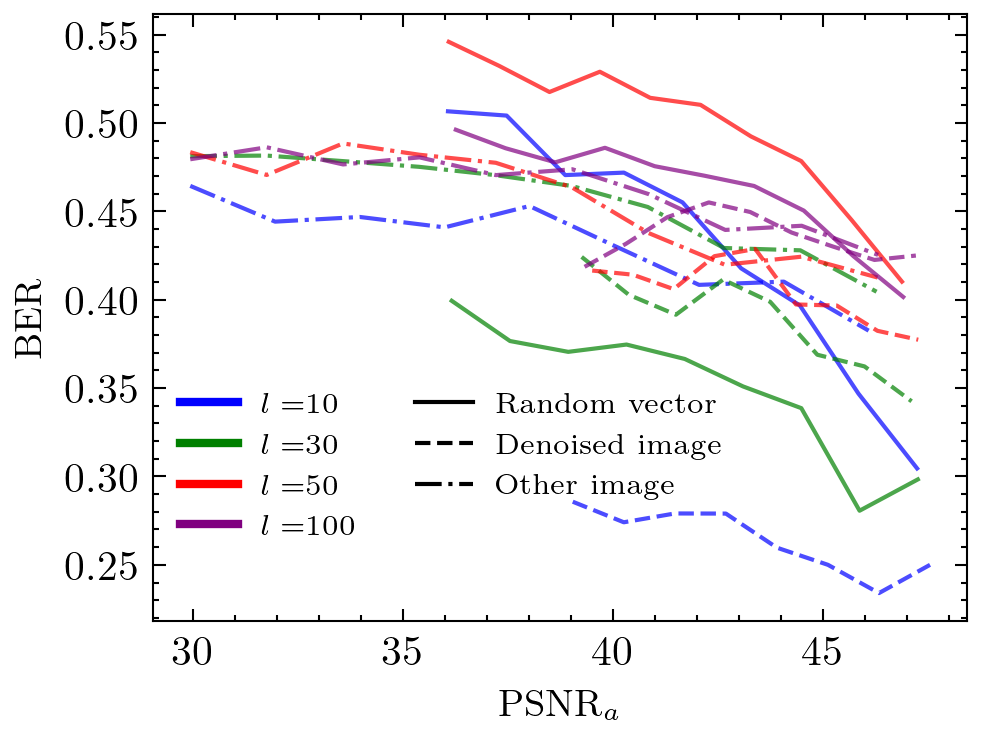}
    \caption{Bit Error Rate for multi-bit watermarking under targeted removal attack with different target image selection strategies. The best results correspond to BER=0.5 (random chance).}
    \label{fig:removal_attack_multi_bit}
\end{figure}

As for multi-bit watermarking, the BER evaluates the success of the attack. The watermark payload is fixed at $\ell\in\{10, 30, 50, 100\}$ bits. The results in Fig.~\ref{fig:removal_attack_multi_bit} demonstrate how the BER depends on the $\text{PSNR}_a$ of the attacked image. The removal efficiency decreases with the increase of $\text{PSNR}_a$. 

The choice of target in the targeted removal attack dictates the different attack efficiency in terms of effective $\text{PSNR}_a$ and achievable BER for different watermark message lengths. The ``other image'' target selection requires largest $\text{PSNR}_a$, i.e., highest possible distortions, to maximally damage the watermarked message for the range of 30-37~dB. The random vector subset space target allows achieves similar values of BER starting at 37~dB but with considerably higher variability of BER values for different message lengths. Finally, the ``denoised image'' selection as a target for the considered removal attack achieves similar results starting from 39~dB under the same impact of message length on BER variability. The overall increase of $\text{PSNR}_a$ leads to the decrease of BER due to the reduction of allowable distortion budget.

One can observe that under the untargeted attacks, the results are somewhat unstable under different $\text{PSNR}_a$. We argue that this is due to the nature of untargeted attacks. Unlike targeted attacks, which push the image latent representation to be as close as possible to the selected target latent representation, the untargeted attacks push the attacked image latent representation far from the watermarked image (cosine similarity between representations is 0). Thus, it can result in an infinite number of optimal solutions.
\section{Conclusion}
This paper investigates the efficacy of copy and removal attacks against a watermarking technique based on the foundation model's latent space.
The results demonstrate that the effectiveness of these attacks increases with the level of adversarial distortions applied. Among the two types of attacks, removal attacks have proven to be more efficient against both watermarking schemes. Copy attacks are relatively easier to perform on zero-bit watermarking.
This is attributed to the more complex nature of multi-bit watermarking latent space spanning.

It is important to note that all experimental results were obtained using the DINOv1 model.
This demonstrates its high vulnerability attacks, and its use for watermarking is not recommended.
Consequently, a future research direction involves investigating a broader class of foundation and autoencoder models in the context of digital watermarking, as well as comparison with classical schemes like Broken Arrows \cite{furon2008broken}. This would help determine whether such vulnerabilities are specific to certain types or consistent across different models.
The latter case implies that watermarking is a specific downstream task that cannot be solved with a public foundation model. 

\bibliographystyle{IEEEtran} 
\bibliography{bib/references-wifs2024}

\end{document}